\documentclass[runningheads]{llncs}
\usepackage[T1]{fontenc}
\usepackage{graphicx}
\usepackage{amsmath}
\usepackage{cite}
\usepackage{xcolor}
\usepackage{booktabs}
\usepackage{float}
\usepackage{caption}
\usepackage{tikz}
\usepackage{url}
\usetikzlibrary{positioning, fit, decorations.pathreplacing}
\begin{document}
\title{Look What You Made Us Cluster: Hate Narrative Extraction from Reddit Discourse}
\titlerunning{Hate Narrative Extraction from Reddit Discourse}
%
\author{Annabelle K. L. Chua\inst{1} \and
Forster J. Khoo\inst{1} \and
Joel C. R. Tan\inst{1} \and
Huey Ting Ang\inst{1} \and
Kheng Hwee Tan\inst{1} \and Joel Y. A. Sim\inst{2} \and Shirley W. H. Ow\inst{2} \and Ria Mundhra\inst{2} \and Elsie C. K. Toh\inst{2} \and Youfeng Xu \and Lynnette H. X. Ng\inst{2}\thanks{The writing and the views of the paper are solely the author's own}}
\authorrunning{Chua et al.}
%
\institute{DSO National Laboratories, 14 Science Park 118226, Singapore \and
Defence Science and Technology Agency, 1 Depot Road 109679, Singapore}
\maketitle              
\begin{abstract}

Narrative extraction allows us to identify online hate narratives, supporting the construction of rigorous detection systems. Existing computational approaches, however, are limited in precision as they rely on semantic representations, which tend to capture only surface-level meaning. To detect more precise and interpretable narratives, we present an extraction pipeline that represents narratives as entity–evaluation pairs. Narratives are extracted using a Large Language Model (LLM) reasoning process that extends Aspect-Based Sentiment Analysis, identifying the aspect, classifying its judgement type as the basis for evaluation, and deriving the evaluation accordingly. Extracted narratives are then clustered using Leiden, following which clusters are resolved to an intended level of granularity through an LLM-guided refinement process. We illustrate this narrative pipeline with English Reddit comments from 2024 that criticize Taylor Swift, analysing a representative cluster that exhibits hate speech patterns to demonstrate its interpretive value.

\end{abstract}

\section{Introduction}

Narrative detection can serve as an important tool for understanding how online discourse spreads and shapes attitudes over time, with particular relevance to online hate, which can emerge from sustained narratives that accumulate gradually across seemingly harmless posts \cite{almagro2026hate}. For instance, narrative similarity has helped identify coordinated user networks, with the disseminated narratives examined to distinguish the agendas of different user groups \cite{ng2023hear}. More broadly, this illustrates the potential of narrative detection to reveal patterns of harmful online activity, since hate narratives can similarly accumulate and spread through networks of users. However, many existing detection methods built on semantic similarity tend to capture only surface-level patterns and may struggle to identify narratives that rely on indirect or implicit language. We therefore propose a framework that uses LLMs to extract and aggregate online narratives. This framework can help capture how narratives shape audience perceptions of social entities, supporting more effective platform governance.


Narratives are characterizations of individuals, groups, or organizations. When clustered into broader patterns, narratives form meta-narratives: overarching storylines that reflect underlying assumptions and beliefs \cite{tannen2008close, ang2017chinese, moss2021everybody}. Building on this, we introduce our narrative extraction pipeline, which identifies entity-specific narratives by combining aspect-based sentiment analysis with judgement frame analysis. These narratives are then clustered using graph-based community detection and structured LLM reasoning to surface a higher-level landscape of meta-narratives. We demonstrate this pipeline on celebrity discourse, given its high volume and the associated risks of online bullying and reputational harm. Specifically, we apply it to English Reddit comments from 2024 discussing Taylor Swift. Beyond this application, the pipeline can be adapted to other social categories where online hate detection might be a concern.

\section{Background}

\subsubsection{Entity Extraction and Resolution.}
Entity extraction and resolution are typically addressed by separate models, from statistical and deep learning models for extraction~\cite{keraghel2024ner} to similarity-based matching, blocking, and clustering for
resolution~\cite{christophides2020overview}. Recent advances show that LLMs can perform extraction directly, achieving performance comparable to fully supervised baselines and outperforming them in low-resource and few-shot settings~\cite{wang2023gptner}. This motivates our investigation into whether a single LLM can also perform resolution reliably, and do so jointly with extraction.

\subsubsection{Narrative Extraction.}
Existing computational approaches to narrative analysis often focus on identifying what narratives exist, without considering how they are framed, limiting the precision of narrative extraction \cite{alieva2024russian, trimmingham2022uyghur}. Such approaches typically rely on semantic representations that capture surface-level similarity, which can produce outputs that are difficult to interpret \cite{ma-etal-2025-cast}. Previous work has attempted to address this by adding additional structure to clusters, such as classifying narratives into predefined story types \cite{ng2021coronavirus}. However, the reliance on manual annotation limits the scalability of such approaches. By deriving discourse structure from theory rather than inferring it after clustering, we extract narratives more precisely and interpretably.

\section{Methodology}

\subsection{Pipeline Overview}
\label{sec:method-pipeline}
\begin{figure}[h]
    \centering   
    \includegraphics[width=0.9\linewidth]{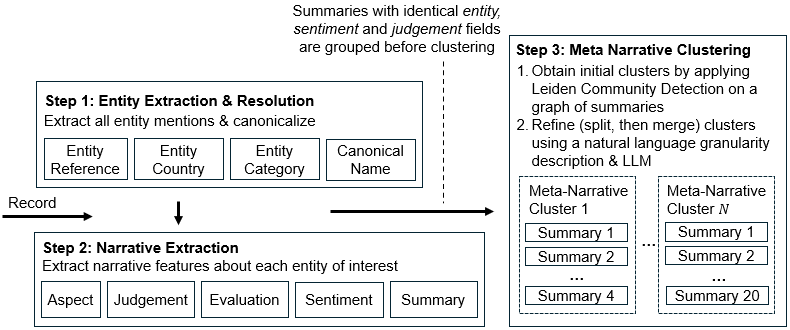}
    \caption{Narrative Pipeline Logic.}
    \label{fig:pipeline-logic}
\end{figure}

Figure~\ref{fig:pipeline-logic} illustrates our three-stage pipeline. Given an input document, we identify entities and resolve them to their canonical forms (Section~\ref{sec:method-entity-extraction}). For each resolved entity, we then extract narrative features that characterize how the entity is portrayed within the document (Section~\ref{sec:method-narrative-extraction}). Finally, these features are grouped into buckets and clustered using community detection. The resulting clusters are then refined with an LLM to produce coherent meta-narratives (Section~\ref{sec:method-meta-narr-extraction}). We applied our pipeline to the ``Exorde Social Media One Month 2024'' dataset~\cite{exorde2024dataset}, analysing English Reddit comments in the Entertainment category. We focused our analyses on ``Taylor Swift'', one of the top extracted entities. String matching on ``Taylor'' or ``Swift'' yielded 13,812 candidate records.



\subsection{Entity Recognition and Resolution}
\label{sec:method-entity-extraction}
Each comment is passed to an LLM, which extracts the list of entities it contains using a single prompt. For each entity, the LLM records its reference (the form in which it appears in the text) and determines its country of residence, along with a category (e.g., individual, group, or organisation), which together support resolving the entity to its canonical form. Resolving entities to a canonical form reduces fragmentation from variations in naming, abbreviations, and indirect references, forming coherent entity groups for downstream narrative analysis. Following this process, the 13,812 candidate records identified in Section~\ref{sec:method-pipeline} were filtered down to 6,715.

\subsection{Narrative Extraction} 
\label{sec:method-narrative-extraction}

Narratives are centered on entities—individuals, groups, and organizations—whose actions, attributes, or experiences form the subject of the story. Following Page's \cite{page2010reexamining} approach to identifying narratives in short online content, each narrative is defined by two key components: an \textit{entity} and an \textit{evaluation} expressed towards that entity. Evaluations capture attitudes conveyed through praise, criticism, or condemnation \cite{hansson2022discursive, oteiza2017appraisal, martin2005language}, shaping audience perception of the entities being discussed \cite{page2010reexamining, desaintlaurent2021memes}. 

Evaluations are further characterized by their \textit{judgement type} \cite{hansson2022discursive, hansson2024blaming}, which is the basis on which the evaluation is made. We identify ethics (sincerity, honesty, or morality) and capability (skill or dependability) as two dominant judgement types for publicly known entities, aligning with the warmth and competence dimensions of social perception identified by the Stereotype Content Model \cite{fiske2002model, cuddy2008warmth}. We also include appearance as a third judgement type, as celebrity and public-figure discourse is heavily shaped by mass-mediated imagery and visual appeal, and appearance-based online hate remains a prevalent yet relatively understudied harm \cite{grasso2024bodyshaming}. A breakdown of judgement type definitions provided to the LLM in the narrative extraction prompt is presented in Table~\ref{tab:judgement-types}.

\begin{table}[h]
\centering
\caption{Judgement Type Definitions in Our Narrative Extraction Prompt}
\label{tab:judgement-types}
\begin{tabular}{@{}p{3cm}p{9cm}@{}}
\toprule
\textbf{Judgement Type}& \textbf{Definition} \\
\midrule
Ethics& Judgements about morality, integrity, honesty, sincerity, kindness, hypocrisy, humility, abuse of power or trust, or harm to others. Some examples include being genuine, dishonest, having double standards, lacking integrity, insincerity.\\
Capability& Judgements about competence, skill, intelligence, knowledge, ability to deliver, or producing high quality work. This would include characteristics related to the job of the person, e.g., a musician/singer who produces good quality music.\\
Appearance& Judgements which are about looks and physical appearance. Examples include being beautiful, sexy, handsome, unattractive.\\
\bottomrule
\end{tabular}
\end{table}

Narrative Extraction is performed by prompting an LLM (Qwen3-30B-A3B-Instruct-2507-FP8) with a document-entity pair (entity obtained from Section \ref{sec:method-entity-extraction}) and a small set of few-shot examples. In a single generation, the model performs a multi-step reasoning procedure that extends Aspect-Based Sentiment Analysis (ABSA) \cite{hua2024absa} to uncover the evaluation expressed towards the entity.

In the first step, the model identifies the aspect—defined as a specific attribute or characteristic of the entity discussed (e.g., leadership). Next, it determines the type of judgement (ethics, capability or appearance) being expressed, if any. This step serves as a useful checkpoint: the presence of a judgement signals that an evaluation is likely being made, reducing both false positives from non-judgemental descriptions and false negatives from subtle, implicit evaluations based on a judgement type. Finally, the LLM assesses the overall evaluation made towards the entity, as well as a sentiment score (on a scale of -2 to 2, for more resolution when monitoring narratives; collapsed to discrete labels of positive, neutral, or negative for bucketing before clustering) reflecting its direction. Thereafter, the LLM condenses the entity, aspect, and evaluation into a single summary field, which can be used for downstream clustering and analyses. The features extracted at this stage are summarized in Table \ref{tab:narrative-features}, and an illustrative example of the pipeline outputs is given below.

\begin{table}[h]
\centering
\caption{Summary of Extracted Outputs in Narrative Detection}
\label{tab:narrative-features}
\begin{tabular}{@{}p{3cm}p{9cm}@{}}
\toprule
\textbf{Feature} & \textbf{Definition} \\
\midrule
Aspect      & The specific characteristic of the entity that is being discussed \\
Judgement   & The underlying basis on which an evaluation is made \\
Evaluation  & Attitudes towards the entity's behaviour or actions \\
Sentiment   & Sentiment of evaluation towards the entity \\
Summary     & Summarize the evaluation relating to the entity's aspect \\
\bottomrule
\end{tabular}
\end{table}

\paragraph{Example Output.} The following comment serves as an input to the narrative extraction pipeline.
\begin{quote}
\textit{``Sick of hearing about Taylor Swift and all the lame music coming out.''}
\end{quote} 
\noindent The pipeline then produces (1) the aspect, \textbf{music quality}; (2) the judgement type, \textbf{capability}; (3) the evaluation, \textbf{lame music coming out}; (4) a \textbf{negative sentiment} toward Taylor Swift; and (5) a summary, \textbf{Taylor Swift's music is criticised as lame and unimpressive}.

\subsection{Meta-Narrative Extraction}
\label{sec:method-meta-narr-extraction}
Semantic clustering approaches such as BERTopic group documents by embedding similarity alone~\cite{grootendorst2022bertopic}. This is inadequate for narrative clustering, as each cluster must be coherent in both the entity it concerns and the sentiment expressed towards it. We address this by grouping summaries from Section \ref{sec:method-narrative-extraction} into buckets sharing a common entity and sentiment polarity. Summaries within each bucket are embedded and clustered using the Leiden algorithm~\cite{traag2019leiden}, which provides an efficient and scalable approach to community detection for large-scale datasets. Clusters are then labelled and refined through the semantic labelling and cluster refinement stages described below.

\subsubsection{Semantic Labelling}

For each cluster, up to 100 centroid-nearest texts are sampled, providing sufficient
representation while remaining within practical LLM input limits. These are passed
to an LLM to produce a cluster title and summary, and the title is embedded to obtain a cluster-level embedding used in downstream refinement.

\subsubsection{Cluster Refinement}

Leiden yields structurally sound clusters, but two issues may persist: cluster titles may be near-identical when the underlying texts describe similar opinions, or conversely, overly general when a cluster comprises texts spanning distinct subtopics. We address both through a refinement stage, where an LLM determines whether a cluster should be split, if its title is overly general, or merged with another, if their titles are overly similar. Both decisions are governed by a user-supplied natural-language granularity description, ensuring that splitting and merging operate at the intended level of detail.

\emph{Cluster Splitting.} Clusters with at least six texts become candidates for this phase. We pass these candidates' title, summary, and a sampled subset of their texts to an LLM alongside the granularity description, to determine whether the cluster is overly general, covering multiple distinct subtopics. Clusters identified as such are sub-clustered using Leiden at a resolution of $2.0$, above the default resolution, to obtain finer-grained clusters which are then relabelled. 

\emph{Cluster Merging.} We construct a cluster similarity graph where nodes are clusters and edges connect pairs with cosine similarity $\geq 0.75$, computed over cluster-level embeddings, comparing each cluster against its 3 nearest neighbors to limit inference cost. Each candidate edge is passed to the LLM with the granularity description, which determines whether the cluster pair is sufficiently similar for merging. Edges deemed sufficiently similar are retained and the rest pruned. Connected components of the resulting graph define merge groups, which are then relabelled in a single LLM call. Pruned edges with cosine similarity $\geq 0.80$ undergo a \emph{relabel} call that produces disambiguating titles and summaries, or leaves the pair unchanged if labels are already sufficiently distinct.

\section{Results and Discussion}

\subsection{Performance of Entity Extraction and Resolution}
\label{sec:method-ent-validation}
A dataset to evaluate the performance of the entity extraction and narrative extraction stages was built by manually coding a total of 90 outputs randomly sampled from the Taylor Swift entity bin, with 30 outputs selected for each judgment type to ensure the reliability of the results. Out of these 90 outputs, mentions of Taylor Swift were accurately identified and assigned to the canonical bin in 89 instances for an accuracy of 98.9\%.

\subsection{Performance of Narrative Extraction }
\label{sec:method-validation}
The 90 sampled outputs in the evaluation dataset from Section \ref{sec:method-ent-validation} were then annotated by three human raters, all of whom were trained using a common coding scheme. Two independent human raters coded the dataset separately. A third rater then resolved disagreements between the two coders to establish a human ground-truth dataset against which the LLM's ratings were compared. The validation was conducted sequentially to mirror the dependency structure of the pipeline, where each feature was validated only when its predecessor was correctly identified (Figure~\ref{fig:pipeline-logic}). For example, sentiment was only assessed within correctly identified evaluations. This ensures that at each stage the LLM and the human rater are assessing the same narrative, and feature-level metrics should be interpreted accordingly as conditioned on correct upstream identification.

\begin{table}[h]
\centering
\caption{Results for Entity, Aspect and Evaluation.}
\label{tab:validation}
\begin{tabular}{ccc@{}}
\toprule
 \textbf{Entity (\%)} & \textbf{Aspect within accurate} & \textbf{Evaluation within accurate} \\
                                             & \textbf{entities (\%)}           & \textbf{aspect and judgement (\%)} \\
\midrule
 98.9& 75.0& 92.5\\
\bottomrule
\end{tabular}
\end{table}

Subjective categorical variables (e.g., evaluation, aspect) were evaluated differently from fixed-category variables (e.g., judgement, sentiment). For the former, we treat disagreements between coders as natural variation in human interpretation rather than error \cite{pavlick2019inherent}, instead assessing whether the LLM's output constituted a defensible, textually supported interpretation. Where disagreement remained, a third coder made the final call. Fixed-category variables, by contrast, were independently coded by annotators and compared directly against the LLM's outputs.

Within correctly identified entities (N = 89), aspect had an accuracy of 75\% (Table~\ref{tab:validation}). Within accurately identified aspects and judgements (N = 65), evaluation accuracy was 92.5\%. It should be noted that feature-level metrics are computed on a shrinking denominator at each stage, which presents an optimistic picture of accuracy in isolation. To provide a holistic estimate of overall performance, we also report end-to-end accuracy by counting narratives where all features are correctly identified: of 90 narratives extracted across our sample of entities, 68.9\% were fully accurately extracted.  

For judgement type, the LLM demonstrated good classification performance, achieving a macro-averaged F1 score of $1.0$ across the three judgement type categories. Interrater reliability for sentiment ratings was assessed with a two-way absolute agreement, single-measure intraclass correlation coefficient. Results indicated excellent reliability between the LLM and ground truth, $\text{ICC} = .93$, 95\% CI $[.89, .96]$, $p < .001$.

While the pipeline demonstrates reasonable performance, the presence of erroneous features suggests that conclusions drawn from individual narrative extractions should be interpreted cautiously. We anticipate that genuinely recurring narratives will consolidate into meaningful meta-narratives, while erroneous narratives, being idiosyncratic, are unlikely to cluster consistently enough to form a meta-narrative. The metrics reported here should therefore be interpreted as a preliminary performance benchmark rather than a definitive constraint on the pipeline's utility.

\subsection{Performance of Meta-Narrative Extraction}

\begin{figure}[h]
    \centering    
    \includegraphics[width=\linewidth]{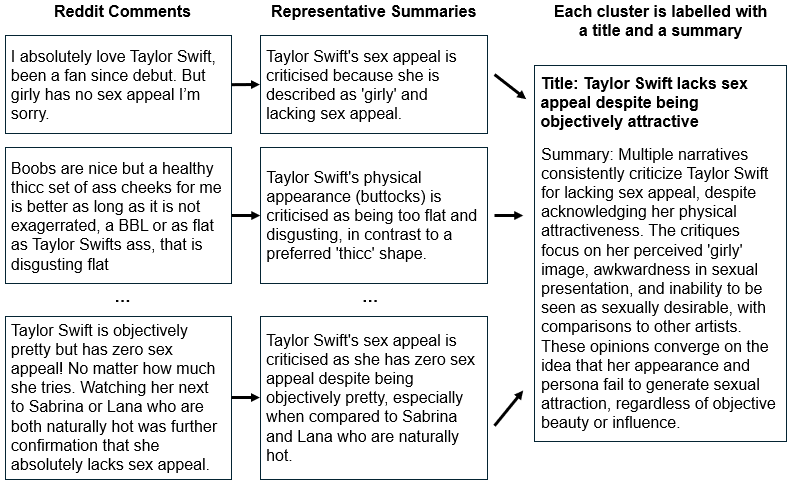}
    \caption{Sample Meta-Narrative Illustration.}
    \label{fig:cluster-example}
\end{figure}

Figure~\ref{fig:cluster-example} presents a meta-narrative recovered by the
pipeline. The example illustrates the pipeline's ability to recover semantically coherent opinion clusters. Although the constituent comments draw on distinct points of comparison---her ``girly'' image, her body shape, and comparisons to other artists---they converge on a shared narrative that Taylor lacks sex appeal despite being objectively attractive. This offers an appropriately granular view of how she is perceived, neither fragmenting into narrow sub-areas such as commentary on a single physical feature, nor collapsing into broad, undifferentiated criticism of her appearance or character. Such thematic specificity could be lost in a broader semantic cluster spanning multiple entities or sentiment polarities. Individual comments in this cluster often read as superficially positive or neutral, prefacing their critique with a compliment, as in ``I absolutely love Taylor Swift...'' or ``Taylor Swift is objectively pretty...''. Evaluated on their surface sentiment alone, such comments might not be flagged by post-level hate detection techniques that assess comments in isolation. It is only once our pipeline extracts the underlying evaluation embedded within each comment, and aggregates it across the cluster, that the broader pattern of characterization becomes visible, along with its potential to shape perception and cause reputational harm over time.


\section{Conclusions and Future Work}
This paper presented a pipeline for narrative extraction and clustering that facilitates more precise and interpretable narratives. This then provides a useful avenue for online narrative monitoring, supporting efforts to detect and understand hate narratives as they emerge.

Several limitations suggest directions for future work. First, clustering quality is
bounded by the accuracy of upstream entity and sentiment extraction. Errors at this
stage propagate into bucket composition and cannot be recovered downstream. Second, the labelling and refinement stages are coupled to the behaviour of the underlying LLM. Both model sensitivity and the non-determinism inherent in LLM calls introduce variation in split and merge decisions, and future work should explore guardrails such as confidence thresholds or ensemble voting across runs to improve consistency. Third, the pipeline processes a static corpus, and extending it to support incremental clustering over evolving document streams would be a natural step toward studying long-term opinion dynamics and coordinated narrative activity in social discourse. Finally, there is a need to fully operationalise the detection of hate narratives by defining criteria at the cluster level. Possible methods include identifying clusters with the lowest average sentiment, or identifying cluster labels with concerning semantic indicators.

%
%
%
\bibliographystyle{splncs04}
\bibliography{references}
\end{document}